\documentclass{llncs}
\usepackage{cite}
\usepackage{graphicx}
\usepackage[cmex10]{amsmath}
\usepackage{algorithmic,color}
\usepackage{stfloats}
\usepackage{multirow}
\usepackage{url}
\usepackage{threeparttable}
\usepackage{multirow}

\title{4D Cardiac Ultrasound Standard Plane Location by Spatial-Temporal Correlation}

\author{
Yun~Gu\inst{1} \and Guang-Zhong Yang\inst{1,2} \and Jie Yang\inst{1,3} \and Kun Sun\inst{4}
}
\institute{
    School of Biomedical Engineering, Shanghai Jiao Tong University, China\and
    The Hamlyn Centre for Robotic Surgery, Imperial College of London, UK \and
    Institute of Image Processing and Pattern Recognition, \\ Shanghai Jiao Tong University, China \and
    Shanghai Xinhua Hospital, China
}

\begin{document}

\maketitle

\begin{abstract}
Echocardiography plays an important part in diagnostic aid in cardiac diseases. A critical step in echocardiography-aided diagnosis is to extract the standard planes since they tend to provide promising views to present different structures that are benefit to diagnosis. To this end, this paper proposes a spatial-temporal embedding framework to extract the standard view planes from 4D STIC (spatial-temporal image correlation) volumes. The proposed method is comprised of three stages, the frame smoothing, spatial-temporal embedding and final classification. In first stage, an $L_0$ smoothing filter is used to preprocess the frames that removes the noise and preserves the boundary. Then a compact representation is learned via embedding spatial and temporal features into a latent space in the supervised scheme considering both standard plane information and diagnosis result. In last stage, the learned features are fed into support vector machine to identify the standard plane. We evaluate the proposed method on a 4D STIC volume dataset with 92 normal cases and 93 abnormal cases in three standard planes. It demonstrates that our method outperforms the baselines in both classification accuracy and computational efficiency.
\end{abstract}

\section{Introduction}\label{sec::introduction}
Echocardiography plays an important part in diagnostic aid in cardiac diseases. It relies on the use of ultrasound images to reflect the information about the heart of a patient. In the process of diagnosis, images are captured from specific views of heart, which are commonly named \textit{heart views} or \textit{standard planes}. Standard planes tend to provide promising views to present different heart structures that are benefit to diagnosis. The task for standard plane location (SPL) exactly refers to selecting 2D planes from a candidate list considering both static and dynamic characteristics from different views of heart.

Automatic recognition of standard planes have been extensively studied in recent works. Most of them focus on ultrasound images with 2D static images, 3D static volumes and 2D-Temporal sequences. For 2D static images, Ebadollahi et.al~\cite{ebadollahi2004automatic} deal with view classification of echocardiograms based on spatial layout of 2D static frames. Ni et.al~\cite{ni2013selective} propose an approach by applying local detectors sequentially on the pre-selected locations with selective search. For 3D volumes, Lu et.al~\cite{lu2008autompr} proposed an automated supervised learning method to detect standard multiplanar reformatted planes (MPRs) from a 3D echocardiographic volume with coarse-to-fine strategy. Chykeyuk et.al~\cite{chykeyuk2013class} build a class-specific random forest regression model for standard plane classification based on 3D static volumes. For 2D-Temporal sequences, Qian et.al~\cite{qian2013synergy} employed bag of visual words based on spatial-temporal features. Sparse coding and max pooling are used in a codebook to generate the representation. 

Previous works usually regard the standard plane location as a classification problem based on individual frames or a set of sequences. In this paper, the data used in our work are 4D STIC volumes where all slices have been captured including both 3D position and temporal information. The task for standard plane location is to select a specific or small set of 2D planes from 4D volumes according to different preferences of views. Different from the previous works, the SPL task on 4D image introduces the following problems:

\begin{enumerate}

\item Restricted by the facility performance, the image quality of 4D STIC volume is often poor. The spatial plane is generated by interpolation and the noise is inevitably introduced.
\item The standard plane of abnormal cases can be quite similar to the plane which is not the standard plane in normal cases.
\item Given the 4D STIC data, the searching space for \textbf{2D plane} is considerably large. The real-time location is in urgent demand.

\end{enumerate}

Considering the issues above, we propose a novel framework for SPL task in 4D STIC volume.
\begin{itemize}
\item Different from the disease diagnosis, it is observed that the SPL tasks focus more on the boundary characteristic but not the fine-grained information. Therefore, we use an $L_0$ smoothing algorithm~\cite{xu2011image} which removes the noise and preserves the boundary simultaneously (For Problem 1).
\item Static SIFT~\cite{lowe2004distinctive} and Spatial-Temporal SIFT (STIP)~\cite{laptev2005space} are deployed to model the feature of standard plane (For Problem 2).
\item Based on SIFT and STIP, we propose a supervised method to learn the discriminative subspace and to generate the compact feature which can obviously reduce the searching consumption (For Problem 2\&3).
\end{itemize}

\section{Methodology}\label{sec::method}
\begin{figure}[!t]
\centering
\includegraphics[width=1.0\textwidth]{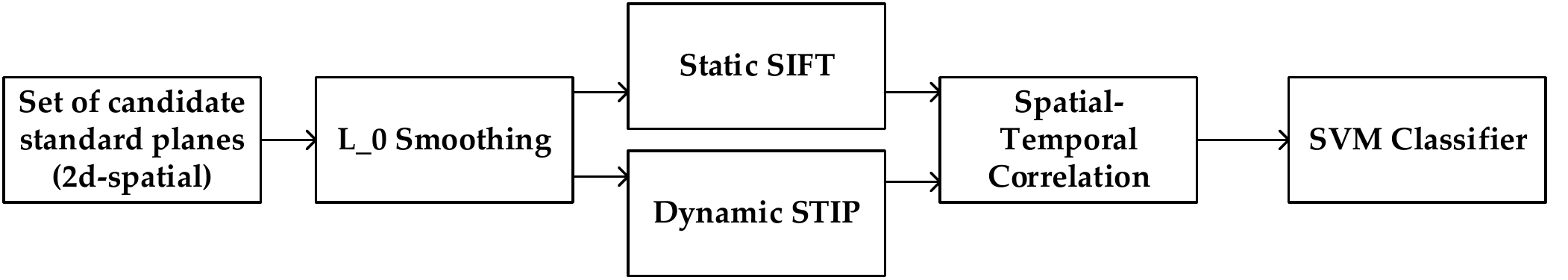}
\caption{Main Framework of the proposed method}\label{fig::framework}
\end{figure}

The main framework of the proposed method is shown in Fig~\ref{fig::framework}. Given the 4D STIC volume data, a large set of candidate 2D planes are generated to identify the standard plane. Each 2D plane is  featured with  a sequence of grey-scale images indicating the cardiac motion. An $L_0$ smoothing filter is firstly deployed to remove the noise in each image. SIFT and STIP features are extracted to capture both static and dynamic information. In order to learn compact representation, a low-dimensional subspace is obtained by embedding SIFT and STIP features via supervised semantic correlation. Finally a support vector machine~(SVM) classifies the candidates into different views.

\subsection{Contour-preserved Smoothing}
\begin{figure}[!ht]
\centering
\includegraphics[width=0.8\textwidth]{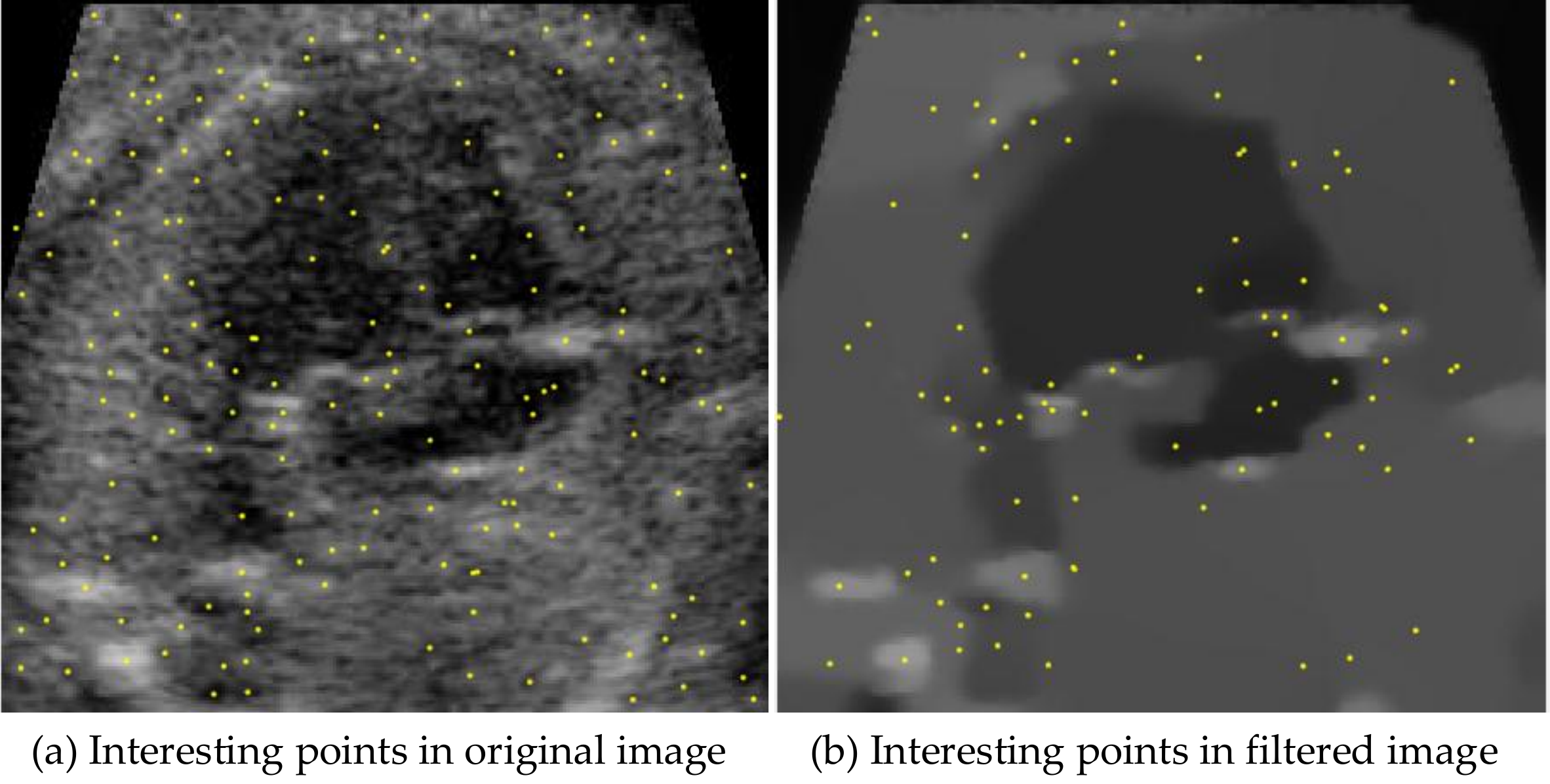}
\caption{Interesting points comparison where interesting points are marked in yellow.}\label{fig::imageVis}
\end{figure}

We deploy the fast $L_0$ smoothing algorithm~\cite{xu2011image} to generate the filtered image and detect the SIFT points as shown in Figure~\ref{fig::imageVis}.

As shown in Figure~\ref{fig::imageVis}, the boundary of original frame is perfectly preserved while the noise is largely reduced in the filter image. We can also find the SIFT points detected in the filtered image mostly distributed along the boundary and the number of points is much smaller than the number in the original image.  As mentioned above, it is observed that the contour is more discriminative to locate the standard planes other than fine grained local features.

\subsection{Spatial-Temporal Features}
The 4D STIC volume data naturally involves the spatial-temporal information. In order to present both dynamic and static features, SIFT \cite{lowe2004distinctive} and STIP\cite{laptev2005space} are used in our work. For each frame, we extract SIFT points as static features that are used to indicate local features and invariant to scale and rotation transforms. \cite{qian2013synergy} demonstrates that the dynamic feature (STIP) is powerful in standard plane location which takes both spatial invariant and temporal structure into consideration. However, the classification accuracy of standard plane is not promising enough according to our experiments on real datasets. STIP prefers to select the interesting points based on spatial-temporal structure. Although STIP performs well in action recognition based on video captured in natural scenes, the quality and sampling rate of ultrasound image sequence is poor without preprocessing. Moreover, the STIP feature cannot fully capture the static information which is also important for standard plane location. Therefore, we aim at learning the common space between dynamic and static features.

\subsection{Multi-view Feature Embedding}
Many strategies are designed for learning the representation from multiple features including multiple kernel learning~\cite{gu2015image} , subspace learning~\cite{gu2015cross}, etc. As mentioned in Section~\ref{sec::introduction}, the searching space is considerably large in standard plane location. Therefore, we tend to learn the feature that integrates both dynamic and static information as well as in compact form. In this section, we follow a supervised embedding scheme to generate the subspace feature based on SIFT and STIP.

Given a 2D plane $p_i$ with $n_i$ grey-scale frames $f_{ij},j=1 \ldots n_i$, the plane is represented by bag-of-words feature with SIFT (denoted by $X_i$) and STIP (denoted by $Y_i$). Each plane is labeled with semantic information including plane description $L^{p}_{i}$ and diagnosis result $L^{d}_i$. $L^{p}_{i}$ is a vector with $n_p+1$ elements $L^{p}_{i}=\{l^{p}_{i,1},\ldots, l^{p}_{i,n_p+1}\}$ where $l^{p}_{i,j}=1$ if current plane belongs to $j$th type of standard plane, otherwise, $l^{p}_{i,j}=0$. If current plane is not a standard plane, the last element $l^{p}_{i,n_p+1}=1$. The diagnosis result label $L^{d}_i=\{l^{d}_{i,1},\ldots, l^{d}_{i,n_d+1}\}$ indicates whether current plane supports the diagnosis of specific disease. Following the similar scheme of plane description, the first $n_d$ elements shows the existence of specific disease (i.e. $l^{d}_{i,j}=1$ supports $j$th disease. Otherwise, $l^{d}_{i,j}=0$). For normal cases, the last element is set to be $1$. Based on the labels, the semantic similarity between two planes can be measured as follows:
\begin{equation}\label{eq::semantic_sim}
S_c(p_i,p_j) = \left\{ 
\begin{array}{ll}
    0 & \quad \textrm{if $(L^{p}_{i})^TL^{p}_{j}=0$}\\
    1+(L^{d}_{i})^TL^{d}_{j} &\quad \textrm{if $(L^{p}_{i})^TL^{p}_{j}=1$}\\
\end{array} \right.
\end{equation}
According to the definition of $S_c$, when two samples belong to different types of standard planes, the similarity is zero. However, the diagnosis information is taken into consideration when samples belong to the same standard plane. Based on the pair-wise similarity, the task for multi-view feature embedding is to learn $W_x$ and $W_y$ mapping $X$ and $Y$ into common latent space that maximizes the semantic similarity as defined in Eq.(\ref{eq::semantic_sim}). Therefore, we have the objective function as follows:
\begin{equation}\label{eq::c_loss}
\min_{W_x, W_y}\|\frac{1}{c}(XW_x)(YW_y)^T-S_c\|_F^2
\end{equation}
where $c$ is the length of the compact codes to learn. $X$ and $Y$ are set of features of all training samples. After learning $W_x$ and $W_y$, the original feature $X$ and $Y$ are transformed into $XW_x$ and $YW_y$ which is compact ($c \ll dim(X_i), c \ll dim(Y_i)$ ) and more discriminative than original feature. In order to obtain the solution of Eq.(\ref{eq::c_loss}), we impose orthogonality constraints, i.e. $(XW_x)^T(XW_x)=nI_c$ and $(YW_y)^T(YW_y)=nI_c$, and then expand the original problem into the following form:
\begin{eqnarray}\label{eqn::conver1}
\begin{aligned}
&\|(XW_x)(YW_y)^T-cS_c\|_F^2\\
&=-2c\cdot tr(W_x^TX^TS_cYW_y)+const\\
\end{aligned}
\end{eqnarray}
Problem in Eq.(\ref{eqn::conver1}) with orthogonality constraints is equivalent to a generalized eigenvalue problem that has closed-form solution. 

\subsection{Standard Plane Location}
The embedded feature $Z=XW_x$ or $Z=YW_y$ is finally fed into a support vector machine (SVM) to determine whether current plane belongs to specific standard plane. 
In order to address the problem of nonlinear cases, a histogram intersection kernel (HIK) is adopted in this paper. For each class of standard plane, we use one-vs-rest strategy to train binary SVM classifier. 

\section{Experiment}\label{sec::exp}

\subsection{Experimental Settings}
\begin{figure}[!t]
\centering
\includegraphics[width=1.0\textwidth]{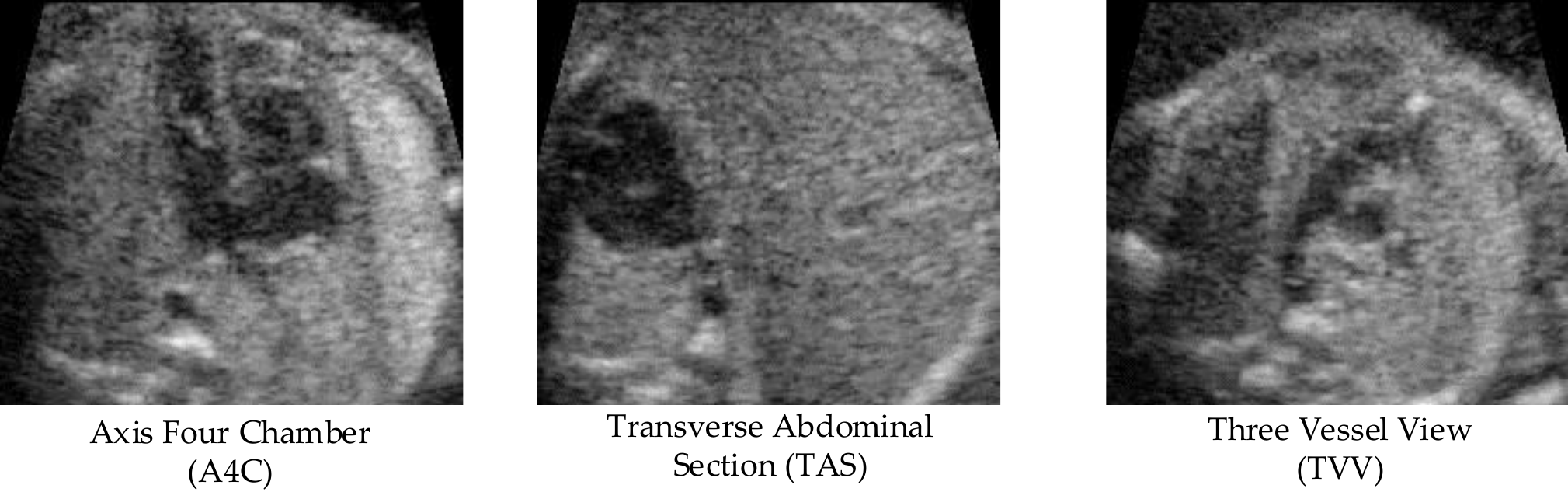}
\caption{Standard Planes}\label{fig::standardplane}
\end{figure}
To evaluate the performance of the proposed embedded feature, a 4D STIC volume dataset is collected by Xinhua Hospital from GE E8 and the data is exported and stored in the DICOM format by 4DVIEW software from GE. The DICOM image has the size of four dimensions with three-dimensional structure and frame numbers ($221\times 164\times 188$ with 40 frames). To simplify the cases of disease, the volume data are divided into two parts with 92 normal cases and 93 abnormal cases. As shown in Figure~\ref{fig::standardplane}, three standard planes including Four Chamber Section (A4C), Three Vessel View (TVV) and Transverse Abdominal Section (TAS) are selected for evaluations. For each volume, we uniformly generate 400 planes as candidate list and, for each standard plane, three candidates are labeled by expert as ground truth. In training step, we randomly select 46 normal cases and 46 abnormal cases from the full dataset. In testing step, we construct two evaluation settings:
\begin{itemize}
\item \textbf{Synthetic data:} For standard planes, we randomly select 75 planes from abnormal cases and 184 from normal cases. For non-standard planes, the numbers of samples from abnormal and normal cases are 1392 and 1304 respectively. Therefore, the evaluation on synthetic data is equivalent to a conventional classification problem. The performance can be directly measured by classification accuracy.
\item \textbf{Real volume:} We regard SPL task as a retrieval problem. For each volume, the query of specific standard plane is conducted by classifying all candidate planes in the same volume. The final performance can be measured by F1-score of standard planes averaged over all volumes.
\end{itemize}

For visual representation, we extract SIFT and STIP interesting points from the sequential image series of planes where the final feature is 5000D Bag-of-Words SIFT and 1000D Bag-of-Words STIP. The dimension of the proposed multi-view feature is 32D. 

Three baseline methods are implemented for comparison:
\begin{itemize}
\item \textbf{Single view feature}. Only SIFT (denoted by "SIFT only") or STIP (denoted by "STIP only") feature is deployed for visual representation for each plane. 
\item \textbf{Direct concatenation}. The SIFT and STIP feature are concatenated into a new vector as visual representation (denoted by "Direct Concate"). The dimension of feature is 6000D.
\item \textbf{Unsupervised CCA}. SIFT and STIP are mapped into a latent space with unsupervised Canonical Correlation Analysis (denoted by "CCA")~\cite{rasiwasia2010new}. The dimension of feature is 32D.
\end{itemize}

All experiments are conducted on a PC with Intel 2.4GHz CPU and 16GB RAM. Methods are implemented with MATLAB. We use LIBSVM package~\cite{chang2011libsvm} for SVM classifier.
\begin{table}[!ht]
    \caption{Classification accuracy on synthetic data}\label{tab::classificationPlane}
    \centering 
    \begin{tabular}{c|c|c|c|c|c}
     \hline
      \hline
     Method   &  STIP only & SIFT only& Direct Concate & CCA&Proposed Method\\
     \hline
      A4C-Abnormal & 0.88 & 0.75 & 0.84 &0.81&0.96\\
      A4C-Normal & 0.78 & 0.87 & 0.83 &0.85&0.87\\
     TAS-Abnormal & 0.83 & 0.21 & 0.87 &0.84&0.96\\
      TAS-Normal & 0.97 & 0.98 & 0.97 &0.92&0.96\\
      TVV-Abnormal & 0.26 & 0.31 & 0.65 &0.56&1.00\\
      TVV-Normal & 0.95 & 0.99 & 0.95 &0.93&0.96\\
     \hline
     \hline
     \end{tabular}
  \end{table}
\subsection{Experimental Results}   

\begin{table}[!ht]
    \caption{F1-score on volume data}\label{tab::classificationVolume}
    \centering
   \begin{tabular}{c|c|c|c|c|c}
     \hline
      \hline
     Method  &  STIP only & SIFT only& Direct Concate & CCA&Proposed Method\\
     \hline
      A4C & 0.38 & 0.44 & 0.42 &0.46&0.53\\
      TAS & 0.34 & 0.13 & 0.34 & 0.17 &0.33\\
      TVV & 0.08 & 0.27 & 0.19 & 0.18 & 0.33\\
     \hline
     \hline
     \end{tabular}
  \end{table}
Table~\ref{tab::classificationPlane} shows the performance of different methods on synthetic data. We can observe from the results that methods using multiple features can achieve better performance than single features. The proposed method can largely improve the performance in all three views where standard planes can be fully detected compared with single feature schemes. Among multi-feature-based methods, the proposed method outperforms both CCA and direct concatenation scheme especially when identifying the standard plane from abnormal cases.  

Table~\ref{tab::classificationVolume} presents the result on real volume data. Similar to the performance on synthetic data, the proposed method achieve high F1-score on A4C retrieval compared with baselines. For more difficult tasks on TVV and TAS retrieval, since SIFT feature cannot fully capture the characteristics in TAS, the fusion of STIP and SIFT does not gain significant improvement where CCA even suffers great loss from SIFT. However, the proposed method is able to obtain close or better performance on recognizing all three views.

\subsection{Computational Analysis}
\begin{table}[!ht]
    \caption{Running Time Comparison}\label{tab::runningTime}
    \centering
   
    \begin{tabular}{c|c|c|c|c|c}
     \hline
      \hline
      Running Time (s)    &  STIP only & SIFT only& Direct Concate & CCA&Proposed Method\\
     \hline
      Train & 0.50 & 5.2 & 6.0 &0.03+2.19&0.03+1.92\\
      Test & 3.35 & 29.04 & 32.50 &0.14&0.14\\
     \hline
     \hline
     \end{tabular}
  \end{table}

Table~\ref{tab::runningTime} shows that the running time of training and testing steps. For CCA and the proposed method, the training part is comprised with the learning of multi-view embedding matrix and SVM parameters. We can observe that the time consumption is much smaller than learning SVM parameters. Since the dimension of the embedded feature is 32D which is much smaller than original feature, the testing time is only 0.14s which is much faster than the rest of approaches while the training time is also acceptable. The real-time detection of standard plane can be implemented. 

\section{Conclusion}
This paper presents a spatial-temporal embedding framework to extract the standard view planes from 4D STIC volumes. We evaluate the proposed method with 92 normal cases and 93 abnormal cases in three standard planes. It demonstrates that our method outperforms the baselines in both classification accuracy and computational efficiency. The future work includes the use of more powerful features (e.g. deep convolutional network) to obtain higher accuracy.

\end{document}